\documentclass[letterpaper, 10 pt, conference]{ieeeconf}  

\IEEEoverridecommandlockouts                              

\usepackage{multicol}
\usepackage[bookmarks=true]{hyperref}
\usepackage{common}
\usepackage{this_doc}
\usepackage{cite}

\usepackage[en-US]{datetime2}
\newcommand{\mydate}[1]{%
  \DTMsavetimestamp{creation}{#1T00:00:00-05:00}%
  \date{\DTMusedate{creation}}%
  {\DTMsetstyle{pdf}%
  \pdfinfo{
    /CreationDate (\DTMuse{creation})
  }}%
}
\mydate{2022-1-28}
\title{SCORE: A Second-Order Conic Initialization for Range-Aided SLAM}
\author{Alan Papalia$^{1,2}$, Joseph Morales$^{1}$, Kevin J. Doherty$^{1,2}$, David M. Rosen$^{3,4}$, John J. Leonard$^{1}$
\thanks{$^{1}$ \CSAIL, \MIT, \MITaddr,~{\tt\footnotesize \{apapalia, jrales, kdoherty, jleonard\}@mit.edu}}
\thanks{$^{2}$ \AOPE, \WHOI, \WHOIaddr}
\thanks{$^{3}$ Department of Electrical \& Computer Engineering, Northeastern University,
Boston, MA 02115, USA, {\tt\footnotesize d.rosen@northeastern.edu}}
\thanks{$^{4}$ Department of Math, Northeastern University,
Boston, MA 02115, USA}
}

\begin{document}

\maketitle
\thispagestyle{empty}
\pagestyle{empty}

\begin{abstract}
    We present a novel initialization technique for the range-aided simultaneous
    localization and mapping (RA-SLAM) problem. In RA-SLAM we consider
    measurements of point-to-point distances in addition to measurements of
    rigid transformations to landmark or pose variables. Standard formulations
    of RA-SLAM approach the problem as non-convex optimization, which requires a
    good initialization to obtain quality results. The initialization technique
    proposed here relaxes the RA-SLAM problem to a convex problem which is then
    solved to determine an initialization for the original, non-convex problem.
    The relaxation is a second-order cone program (SOCP), which is derived from
    a quadratically constrained quadratic program (QCQP) formulation of the
    RA-SLAM problem. As a SOCP, the method is highly scalable. We name this
    relaxation \ScoreFull~(SCORE). To our knowledge, this work represents the
    first convex relaxation for RA-SLAM. We present real-world and simulated
    experiments which show SCORE initialization permits the efficient recovery
    of quality solutions for a variety of challenging single- and
    multi-robot RA-SLAM problems with thousands of poses and range measurements.
\end{abstract}

\section{Introduction}

Range-aided simultaneous localization and mapping (RA-SLAM) is a key robotic
task, with applications in planetary \cite{boroson20iros}, subterranean
\cite{funabiki21ral}, and sub-sea \cite{Newman03icra,Bahr06iser,Bahr12iros}
environments. RA-SLAM combines range measurements (e.g., distances between
acoustic beacons) with measurements of relative rigid transformations (e.g.,
odometry) to estimate a set of robot poses and landmark positions.

RA-SLAM differs from pose-graph simultaneous localization and mapping (SLAM) in
that the sensing models of range measurements induce substantial difficulties
for state-estimation. However, ranging is a valuable sensor modality and
further developments in RA-SLAM could substantially advance the state of robot
navigation.

The state-of-the-art formulation of SLAM problems is as \textit{maximum a
posteriori} (MAP) inference, which takes the form of an optimization problem.
However, because robot orientations are a non-convex set, the MAP problem is
non-convex. Additionally, in RA-SLAM, the measurement models of range sensing
introduce further non-convexities. As a result, standard approaches to solving
these MAP problems use local-search techniques and can only guarantee locally
optimal solutions. Thus, good initializations to the MAP problem are key to
obtaining quality RA-SLAM solutions.

\TitleFigure

This work approaches initialization for the non-convex MAP estimation problem
through \textit{convex relaxation} \cite{boyd04book}, the construction of a
convex optimization problem which attempts to approximate the original problem.
Convex problems can be efficiently solved to global optimality, and many convex
relaxations have shown success as initialization strategies in related robotic
problems
\cite{carlone15icra,carlone15iros,giamou19ral,so07mathematicalprogramming}.

This work presents SCORE as a novel methodology for initializing RA-SLAM
problems. SCORE applies to 2D and 3D scenarios with arbitrary numbers of robots
and landmarks. As SCORE is a SOCP, it is easily implemented in existing
convex optimization libraries, and scales gracefully to large problems. We
summarize our contributions as follows:

\begin{itemize}
      \item A novel, convex-relaxation approach to initializing RA-SLAM
            problems.
      \item The first QCQP formulation of RA-SLAM, connecting
            RA-SLAM to a broader body of work
            \cite{rosen19ijrr,carlone15iros,giamou19ral,so07mathematicalprogramming}.
      \item Our implementation of SCORE is open-sourced\footnote{\url{\RepoURL}}.
\end{itemize}

\section{Related Work}

The current state-of-the-art formulation of RA-SLAM is through nonlinear
least-squares (NLS) optimization based upon MAP inference
\cite{dellaert2017factor}. This work specifically focuses on
\textit{initialization} for this MAP formulation. Though we focus on MAP
estimation, there are many important previous formulations of RA-SLAM using:
extended Kalman filters \cite{Newman03icra, menegatti09icra,
Djugash09iros,djugash09springer}, particle filters \cite{gonzalez09ras,
blanco08icra}, mixture models \cite{blanco08iros}, and NLS optimization
\cite{boroson20iros, funabiki21ral, herranz14icra}.

We first survey previous initialization strategies for RA-SLAM problems and then
expand the scope to cover initialization strategies for general pose-graph SLAM
problems. As our initialization approach is a convex relaxation of the RA-SLAM
problem, we discuss related convex relaxations in robotic state-estimation.
Finally, as the key challenge that differentiates RA-SLAM from pose-graph SLAM
is the nature of range measurements, we discuss the closely related field of
sensor network localization (SNL) and convex relaxations developed specifically
for SNL problems.

\subsection{Initialization in RA-SLAM}

Notable work in the single-robot RA-SLAM case used spectral graph clustering to
initially estimate beacon locations \cite{Olson06joe}. Similarly,
\cite{boots13icml} used spectral decomposition to jointly estimate robot poses
and beacon locations. However, these approaches \cite{Olson06joe,boots13icml} do
not account for sensor noise models and do not readily extend to multi-robot
scenarios. Other initializations for multi-robot RA-SLAM used coordinated
movement patterns to find the (single) relative transform between robots
\cite{guo2017ijmav,li20arxiv}.

Finally, a series of works developed increasingly sophisticated methods to
compute relative transforms between two robots using range and odometry
measurements
\cite{zhou08tro,trawny07iros,trawny10rss,trawny10tro,jiang20itaes,li20iros,Li2022ral}.
While these approaches represent great progress, they are only capable of
solving for a single inter-robot relative transform and rely on (noisy)
odometric pose composition to initialize all other poses. In fact,
\cite{jiang20itaes,li20iros,Li2022ral} utilize convex relaxation to solve the
relative transformation problem.

\subsection{Initialization and Convex Relaxation in State-Estimation}

In pose-graph SLAM many existing initializations solve approximations of the MAP
problem. \cite{carlone15icra} leveraged the relationships between rotation and
translation measurements in SLAM and explored the use of several rotation
averaging algorithms
\cite{martinec07cvpr,govindu01cvpr,fredriksson13lnsc,tron14tac} to obtain
initial estimates. The same relationships were used to approximate the SLAM
problem as two sequential linear estimation problems \cite{carlone14ijrr}.
However, these approaches do not consider range measurements and thus do not
extend to RA-SLAM problems.

With exception to \cite{tron14tac}, these initialization procedures represent
convex relaxations of the pose-graph SLAM problem. Subsequent works in
pose-graph SLAM developed convex relaxations which obtained exact
solutions to the non-convex MAP problem \cite{rosen19ijrr,
carlone15iros,briales17ral,tron15rssworkshop, fan20tro
}. Other works developed exact convex relaxations for
the problems of extrinsic calibration \cite{giamou19ral}, two-view relative pose
estimation \cite{briales18cvpr,garcia-salguero21ivc}, and spline-based
trajectory estimation from range measurements with known beacon locations
\cite{pacholska20ral}. With exception to \cite{pacholska20ral}, these convex
relaxations also do not extend to range measurements. However,
\cite{pacholska20ral} required known beacon locations, allowed only for range
measurements, and only considered single-robot localization. SCORE allows for
multi-robot RA-SLAM, and combines measurements of rigid transformations and
ranges without necessitating any information \textit{a priori}.

A number of convex relaxations exist for sensor network localization (SNL). As
SNL centers around point-to-point distance measurements, these works are closely
related to SCORE. Previous works  developed and analyzed semidefinite
program (SDP) \cite{biswas06tase,
so07mathematicalprogramming, shamsi13dgo} and SOCP \cite{tseng07siam,naddafzadeh-shirazi14twc,doherty01infocom,
srirangarajan08twc}
relaxations of the SNL problem. Whereas these relaxations only
consider range measurements and can only estimate the Euclidean position of
points, SCORE allows for rigid transformation measurements and estimation of
poses. Additionally, as SCORE is a SOCP, it is substantially more scalable than
these SDP relaxations.

\subsection{Novelty of Our Approach}

SCORE is similar to \cite{martinec07cvpr} in its relaxation of rotations and to
\cite{naddafzadeh-shirazi14twc} in its relaxation of range measurements.
However, it differs from these works in that it jointly considers both relative
pose and range measurements. Notably, SCORE generalizes the
chordal-initialization of \cite{martinec07cvpr} and the SOCP relaxation of
\cite{naddafzadeh-shirazi14twc}.

Additionally, there is much commonality to
\cite{jiang20itaes,li20iros,Li2022ral} in that the convex relaxation of SCORE
considers both range and relative transformation measurements. Unlike these
works, SCORE generalizes to arbitrary dimensions (e.g. 2- or 3-D) and
multi-robot cases. Furthermore, SCORE differs in that the convex relaxation is
derived from the full MAP problem, and thus utilizes all measurements and
jointly solves for an initialization of all RA-SLAM variables.

\section{Notation and Mathematical Preliminaries}

\textbf{Symbols:} This work generalizes across dimensionalities (e.g., 2-D or
3-D). As such the mathematical presentation will assume to be working in
$d$-dimensional space.  We denote the $d$-dimensional identity matrix as $I_d$
and the special orthogonal group as $\SOd$. We refer to the set of real values
as $\R$ and the set of nonnegative reals as $\RnonNeg$. Furthermore, we indicate vectors and
scalars with lowercase symbols (e.g., $\tran$) and matrices with uppercase
symbols (e.g.,  $\rot$). Noisy measurements are indicated with a tilde (e.g.,
$\nrot$). The true (latent) value of a quantity will be underlined (e.g.,
$\trot$)

\textbf{Mathematical Operators:} We denote the matrix trace as $\tr (\cdot)$. A
diagonal matrix with diagonal elements $[a_1, \ldots, a_n]$ is indicated as
$\diag ([a_1, \ldots, a_n])$. The vector 2-norm is indicated as $\|
\cdot \|_2$ and the Frobenius norm is indicated as $\| \cdot \|_F$. We
note the following relationship between the Frobenius norm and trace operators:
$\| A \|_F^2 = \tr(A A^\top) = \tr(A^\top A)$.

\textbf{Probability Distributions:} We indicate a Gaussian probability
distribution over Euclidean space, with mean $\mu \in \dvec$ and variance
$\Sigma \in \dmat$ as $\gaussian (\mu, \Sigma)$. Similarly, we denote the
isotropic Langevin probability distribution\IsotropicLangevinFootnote over
matrix elements of the special orthogonal group, $\SOd$, with mode $\rot_\mu \in
\dmat$ and concentration parameter $\kappa \in \R$ as $\langevin (\rot_\mu,
\kappa)$ \cite{chiuso08infosys}.
This is the distribution whose probability density function is:
\begin{equation}
    \label{eq:langevin-distribution}
    \begin{array}{ll}
        \prot \sim \text{Langevin}(\rot_\mu, \kappa) \\ \prob( \prot )
        = \frac{1}{c(\kappa)} \exp\{\kappa \tr (\rot_\mu^\top
        \prot)\}
    \end{array}
\end{equation}
with normalizing constant $c(\kappa)$.

\section{Range-Aided SLAM Formulation}

In this section we model the RA-SLAM problem and derive its
corresponding MAP estimation problem.

\subsection{Inference Over a Graph}

We formulate RA-SLAM as inference over a directed graph, with nodes as variables
(i.e. robot poses and landmark positions) and the edges representing
measurements. Without loss of generality, we assume directionality in the edges.
The edge-set of all measurements of relative rigid transformations is denoted
$\TransformEdges$, where $\dedge \in \TransformEdges$ represents a measured
rigid transformation from node $i$ to node $j$. Similarly, the edge-set of all
range measurements is denoted $\RangeEdges$ and each $\dedge \in \RangeEdges$
represents a measured distance from node $i$ to node $j$.

\subsection{Measurement Noise Models}
\label{sec:noise-models}

For relative rigid transformation measurements we follow the formulation of
\cite{rosen19ijrr} in the noise model we assume over our
measurements. We denote the true translation and orientation of node $i$ as
$\ttran_i \in \dvec$ and $\trot_i \in \SOd$, with the true relative rotation from node $i$ to node
$j$ as $\trot_{ij} \triangleq \trot_{i}^\top \trot_{j}$. Similarly, we denote
the true relative translations as $\ttran_{ij} \triangleq \trot_{i}^\top (
        \ttran_{j} - \ttran_{i} )$. We indicate noisy measurements with a tilde (e.g.,
$\nrot$):
\begin{equation}
    \begin{aligned}
        \ntran_{i j} & =\ttran_{i j}+\ptran_{i j}, & \ptran_{i j} & \sim \mathcal{N}\left(0, \tau_{i j}^{-1} I_{d}\right), \\
        \nrot_{i j}  & =\trot_{i j} \prot_{i j},   & \prot_{i j}  & \sim \langevin \left(I_{d}, \kappa_{i j}\right).
    \end{aligned}
\end{equation}

We assume the following generative Gaussian model for range measurements:
\begin{equation}
    \label{eq:range-measurement-model}
    \begin{aligned}
        \ndist_{i j}   & = \lVert \ttran_i - \ttran_j \rVert_2
        +\pdist_{i j}, & \pdist_{i j}                          & \sim \mathcal{N}\left(0, \sigma_{i j}^2 \right). \\
    \end{aligned}
\end{equation}

\subsection{MAP Estimation for RA-SLAM}

We write the MAP problem corresponding to the sensor models of
\cref{sec:noise-models}. The MAP problem is as follows, where $\tran_i \in
    \dvec$ and $\rot_i \in \SOd$ denote the variables corresponding to the estimated
translation and rotation of node $i$. Bolded symbols indicate groupings (e.g.,
$\textbf{\tran} \triangleq \{\tran_i~\forall~i=1,\ldots,n \}$):
\begin{equation}
    \label{eq:general-MAP}
    \begin{array}{r}
        \max\limits_{ \bf\tran , \bf\rot } ~ \prob ( \bf \ntran , \bf \nrot, \bf \ndist \mid \bf \tran , \bf \rot).
    \end{array}
\end{equation}
This general MAP problem of \cref{eq:general-MAP} is equivalently solved by
minimizing its negative log-likelihood \cite{dellaert2017factor}. Given the
models of \cref{sec:noise-models}, the MAP estimate of the RA-SLAM
problem can be expressed as the following NLS problem:
\begin{equation}
    \label{eq:ra-slam-nls}
    \begin{aligned}
        \min\limits_{\substack{ \tran_i \in \dvec                                                                                    \\ \rot_i \in \SOd}} ~~
          & \sum\limits_{\dedge \in \TransformEdges} \kappa_{ij} \lVert \rot_j - \rot_i \nrot_{ij} \rVert_F^2                        \\
        + & \sum\limits_{\dedge \in \TransformEdges} \tau_{ij} \left \lVert \tran_j - \tran_i - \rot_i \ntran_{ij} \right \rVert_2^2 \\
        + & \sum\limits_{\dedge \in \RangeEdges} \frac{1}{\sigma^2_{ij}} ( \lVert \tran_i - \tran_j \rVert_2 - \ndist_{ij})^2        ~ .
    \end{aligned}
\end{equation}
where $\| \cdot \|_F$ denotes the matrix
Frobenius norm:

In \cref{eq:ra-slam-nls} the first two summands correspond to relative
transformation measurements whereas the third summand corresponds to
range measurements. Derivation of the cost terms in the first two summands can
be found in \cite{rosen19ijrr}. The cost terms in the third summand follow
immediately from the negative log-likelihood of the model in
\cref{eq:range-measurement-model}.

Notice that there are two distinct sources of non-convexity in \cref{eq:ra-slam-nls}. The rotation variables are members of the special orthogonal group,
i.e., $\rot_i \in \SOd$, which is a non-convex set. Additionally, the cost terms
due to range measurements are non-convex due to the $ \lVert \tran_i - \tran_j
    \rVert_2 $ component.

\section{SCORE: Second Order Conic Relaxation}
\label{sec:ra-slam-score}
In this section we reformulate the MAP problem of \cref{eq:ra-slam-nls} as a
QCQP. We then show how to construct SCORE by relaxing this QCQP to a SOCP.

\subsection{RA-SLAM as a QCQP}
\label{sec:ra-slam-qcqp}
 We perform two changes to convert the non-convex NLS problem of
 \cref{eq:ra-slam-nls} problem to the QCQP of \cref{eq:ra-slam-qcqp}. First, we
 rewrite the $\SOd$ constraint as the orthonormality constraint and remove the
 associated determinant constraint.  Secondly, we introduce the auxiliary
 variables, $\dist_{ij} \in \RnonNeg$ to rewrite the range-measurement cost
 terms as quadratics.  In this formulation the cost terms are quadratic and
 convex with respect to the problem variables and all constraints are quadratic
 equality constraints. Thus, \cref{eq:ra-slam-qcqp} is a QCQP.

Previous work \cite{carlone15iros,briales17ral,tron15rssworkshop} demonstrated
that $\det(\rot_i) = 1$ can be written as a set of quadratic equalities and that
the contribution of this determinant constraint is negligible, justifying its
removal. With exception to the determinant constraint, this QCQP is an exact
reformulation of \cref{eq:ra-slam-nls}:
\RaSlamQcqp

\subsection{Relaxing Equation (\ref{eq:ra-slam-qcqp}) to a SOCP}
\label{sec:ra-slam-socp}

We will now introduce the primary contribution of our paper, a \ScoreFull~(SCORE). As
previously mentioned, SCORE naturally arises as a relaxation of the QCQP in
\cref{eq:ra-slam-qcqp}. We emphasize that \cref{eq:ra-slam-qcqp} is not solved
in our approach, as it is not computationally practical to do so.
\cref{eq:ra-slam-qcqp} is important as an intermediate representation to derive
SCORE.

We relax the orthonormality constraints by removing them entirely, eliminating
that specific source of non-convexity. To eliminate the non-convexity from the
quadratic equality constraint on our auxiliary variables ($\dist_{ij}$) we first
recall that $\dist_{ij} \geq 0$, so we can drop the squares on each side of the
inequality without modifying the feasible set. From here, we can relax the
equality constraint to the convex inequality constraint $ \dist_{ij} \geq \lVert
    \tran_i - \tran_j \rVert_2$. Importantly, this specific constraint relating the
auxiliary variables and position variables is a (convex) second-order cone
constraint \cite{boyd04book,alizadeh03mathematicalprogramming}. This relaxed
problem takes the following form, where $\CostFunction$ is the same cost
function as in \cref{eq:ra-slam-qcqp}.
\begin{equation}
    \label{eq:ra-slam-socp}
    \begin{aligned}
         \min\limits_{\bf \tran, \bf \rot, \bf \dist} ~~ & \CostFunction \\
         \st & \dist_{ij} \geq \lVert \tran_i - \tran_j \rVert_2,~\forall \dedge \in \RangeEdges \\
          & \rot_0 = I_d
    \end{aligned}
\end{equation}
With this previously mentioned second-order cone constraint on $\dist_{ij}$, it follows that
\cref{eq:ra-slam-socp} is a SOCP, i.e., it is the minimization of quadratic cost
functions over the intersection of an affine subspace with second-order
cones. This is a known form of SOCPs \cite{alizadeh03mathematicalprogramming},
and thus existing solvers (e.g., \cite{gurobi}) can efficiently solve
\cref{eq:ra-slam-socp} to global optimality. Similarly to \cite{martinec07cvpr},
the first rotation ($\rot_0$) is constrained to prevent a trivial solution of
all zeros.

\section{Using SCORE to Initialize Local-Optimization}
\label{sec:score-to-initialize}

Though SCORE can be solved to global optimality, in general the SCORE solution
will differ from that of the original problem due to the relaxed constraints.
This is not a serious limitation, as the intent is that the SCORE solution will
be close to the solution of the original problem. If true, the SCORE solution
can be projected to the feasible set of the original problem, and thereby serve
as a robust and principled initialization procedure.

Projecting the SCORE solution to the original problem's feasible set only
involves the estimated rotations. As SCORE relaxes the $\SOd$ constraints, the
estimated values, $\rot_i$, will generally not be valid rotation matrices. As in
\cite{martinec07cvpr}, we project the approximate rotation estimates to the
nearest rotation matrix via singular value decomposition, as follows:
\begin{align}
    & \orot_i = \argmin_{\rot_i}~~\lVert \rot_i - \erot_{i,approx} \rVert_F^2 \label{eq:single-rotation-frob} \\
    & U,S,V^\top = \SVD(\erot) \label{eq:SVD} \\
    & \orot_i = U ~ \diag([1, 1, \det(U V^\top)]) ~ V^\top \label{eq:SOd-projection}
\end{align}
Projecting all approximate rotations to $\SOd$ produces a valid set of pose and
landmark estimates. We then use this projected solution to initialize the MAP
problem in \cref{eq:ra-slam-nls} and solve the problem with an iterative,
local-search based NLS solver (e.g., \cite{dellaert2012techreport}) to refine
the SCORE initial estimate.
\section{Experiments}
\label{sec:experiments}

In this section we present an experimental evaluation of SCORE's performance on
several RA-SLAM problems, consisting of (1) three real-world trials
involving an autonomous underwater vehicle (AUV) and acoustic beacons, (2)
simulated single-robot experiments with a single static beacon, and (3)
simulated multi-robot scenarios with inter-robot ranging. These experiments were
chosen to
respectively demonstrate (1)
SCORE's utility in real-world settings, (2) the challenge of RA-SLAM in even
seemingly simple cases (single-robot and single-beacon), and (3) SCORE's
performance in the challenging, but useful, scenario in which the relative
transformation between robots must be estimated from odometry and range
measurements.

\subsection{Experimental Evaluation}
\label{sec:experimental-eval}

We compare SCORE to other initialization techniques using the MAP estimate
obtained from the given initialization. Our SCORE implementation follows the
details of \cref{sec:score-to-initialize}, using Drake \cite{drake} and the
Gurobi convex solver \cite{gurobi}.

We emphasize that the initialization strategies compared against represent
either idealistic scenarios (ground truth values) or the existing
state-of-the-art (odometric approaches) for the problems presented in these
experiments. Whereas sophisticated initialization techniques exist for
pose-graph SLAM problems, RA-SLAM lacks such techniques.

Quantitative analysis was in comparison to ground truth values and was based on
absolute pose error (APE) \cite{grupp2017evo}.

\subsection{Real-World AUV Experiments}

The AUV data was collected using four acoustic beacons and a Bluefin Odyssey III
off the coast of Italy.  As GPS was not available for most of the collection, we
obtained ground truth AUV positioning via MAP estimation, using GPS for the
initial pose and combining odometry and acoustic ranging with known beacon
locations to localize the remaining trajectory. Ground truth beacon positions
were obtained by surveyed acoustic trilateration prior to the experiment. We
obtained odometry measurements by estimating speed from Doppler velocity log
(DVL) measurements and combining the speed estimate with heading from the INS
magnetometer. We extracted range measurements via two-way time-of-flight ranging
with the acoustic beacons.

\SingleRobotGoatsTrajectoryFigure
\SingleRobotGoatAPEBarPlot

\subsubsection{Single-Robot Odometry Initialization (Odom)}
\label{sec:single-robot-odom-init}

The Odom initialization set the first pose to be the identity matrix and used
composition along the (noisy) odometry chain to obtain the remaining poses. The
static beacon initialization uses a random sample drawn from the environment, as
done in standard techniques \cite{Newman03icra,guo2017ijmav}.

\subsubsection{Evaluation of Single-Robot Experiments}

As seen in \cref{fig:single-robot-trajs}, the SCORE initialized solution closely
matched the ground truth trajectory. The APE results in
\cref{fig:single-trajectory-error} support this argument, showing consistently
lesser error for the SCORE initialized solution in comparison to the Odom
solution.

\subsection{Generation of Simulations}
\label{sec:generation-simulated-experiments}

We simulated RA-SLAM problems in which robots moved along a grid and
measurements used the noise models of \cref{sec:noise-models}. In all simulated
trials the initial position of any robots or beacons were randomly placed within
the simulated environment. At each time step odometry measurements were recorded
for each robot and each robot had a 50\% chance of generating a range
measurement. If generated, the remaining endpoint of the range measurement was
chosen by uniform sampling from the set of all other robots and beacons present.

\SingleRobotTrajectoryFigure
\SingleRobotAPEBoxPlot

To evaluate the effect of different sensor noise levels, we tested low- and
high-noise cases for both the single-robot and multi-robot scenarios. Each set
of noise parameters comes from values found in previous real-world experiments.
All odometry measurements were 1-meter movements, so a standard deviation of
0.01 meters in translation measurements corresponds to 1\% of the distance
traveled.

The low-noise setting uses $\sigma^{2}_{ij} = \lowSigmaSq$, $\tau^{-1}_{ij} =
\lowTauInv$, and $\kappa^{-1}_{ij} = \lowKappaInv$. This corresponds to standard
deviations of $\lowSigma$ meters, $\lowTauInvSqrt$ meters, and 0.002 radians
(0.41 degrees) for distance, translation, and rotation measurements,
respectively. The high-noise setting uses $\sigma^{2}_{ij} = \highSigmaSq$,
$\tau^{-1}_{ij} = \highTauInv$, and $\kappa^{-1}_{ij} = \highKappaInv$.

\subsection{Simulated Single-Robot Experiments}
\label{sec:single-robot-experiments}

In this section we evaluate the single-robot experiments, which considered
single robot and single static ranging beacon. For both the high and low noise
cases we generated 20 trials, with each trial consisting of 400 poses. For each
trial we compare the results of three different initialization strategies:
SCORE, simple odometry composition, and GT-Init (in which the ground-truth is
used as initialization).

\subsubsection{Evaluation of Single-Robot Experiments}

As seen in \cref{fig:single-robot-trajs}, the SCORE initialized solution
appeared to exactly match the GT-Init initialized solution. This is supported by
the quantitative results in \cref{fig:single-trajectory-error}, which suggest
that SCORE and GT-Init initialization effectively resulted in the same APE for
the trials run. In contrast to SCORE and GT-Init, the Odom initialization
obtained notably worse solutions (i.e. substantially higher APE and
qualitatively incorrect).  In \cref{fig:single-trajectory-error} it can be seen
that the Odom initialization resulted in an increase in APE of up to two orders
of magnitude.

\subsection{Multi-Robot Experiments}
\label{sec:multi-robot-experiments}

\MultiRobotTrajectoryFourPanelFigure

To test a scenario in which initialization is difficult, we
simulated scenarios with four robots and no ranging beacons. For both the high
and low noise cases we generated 20 trials with 400 poses per robot (1600 poses
per trial). We compare the results of four different initialization strategies:
SCORE, ground truth (GT-Init), odometry with perfect first pose (Odom-P), and
odometry with random first pose (Odom-R). We describe these approaches in detail
in \cref{sec:multi-robot-init}.

\subsubsection{Multi-Robot Initialization}
\label{sec:multi-robot-init}

GT-Init is an idealistic case in which the ground truth state is known \textit{a
priori} and all variables are initialized to their true values. While RA-SLAM is
not needed if the true state is known, GT-Init is meant to serve as an upper
bound for reasonable quality of an initialization technique.

In comparison to the GT-Init initialization, the odometry based initializations
represent more realistic scenarios in which either the initial pose of each
robot is known (Odom-P) or no prior information exists (Odom-R). Odom-P and
Odom-R only differ in choosing the starting pose for each robot. After
determining the starting pose for a given robot both Odom-P and Odom-R compose
odometry measurements to initialize the remaining poses for all robots.

In Odom-P we assume a perfectly known starting pose of each robot and compose
the odometry chain to initialize the other pose variables. Such a situation may
occur in marine robotics, where AUVs begin with GPS before diving underwater.
In contrast to Odom-P, for Odom-R we assume no prior information and choose the
starting pose for each robot randomly. We note that Odom-R is a standard
approach to initialization for multi-robot RA-SLAM \cite{guo2017ijmav,li20arxiv}
and is arguably the fairest comparison for SCORE as both require no prior
information.

\subsubsection{Evaluation of Multi-Robot Experiments}

As seen in the example multi-robot trial displayed in
\cref{fig:multi-robot-trajs}, SCORE provides reliable estimates of the
trajectories of the four robots. In this trial the solution obtained from the
SCORE initialization qualitatively matches the true trajectories. In addition,
the refined SCORE initialization appears to match the trajectories estimated
using the GT-Init initialization and provide a substantially better estimate
than that provided by the Odom-P and Odom-R initializations.

These qualitative observations from \cref{fig:multi-robot-trajs} are supported
by the quantitative analysis of the APE in \cref{fig:multi-trajectory-error}. We
observe that for the high-noise multi-robot experiments the SCORE initialization
strategy resulted in similar APE to the GT-Init initialization and had notably
reduced APE in comparison to Odom-P and Odom-R. For the low-noise multi-robot
experiments SCORE resulted in higher APE than the GT-Init initialization
approach but lesser APE than either of the odometric initialization strategies.

\section{Discussion}

In summary, the results shown in
\cref{sec:experiments} demonstrated
that SCORE works in real-world settings and outperforms state-of-the-art
initialization techniques for a wide range of RA-SLAM problems. We emphasize
the difficulty of the multi-robot problems presented and that SCORE is capable
of providing quality initializations for multi-robot RA-SLAM problems with only
odometry and inter-robot range measurements.

\MultiRobotAPEBoxPlot

Furthermore, in evaluating SCORE's performance on these multi-robot experiments
we emphasize that Odom-R is the only other initialization technique which
assumes the same amount of prior information as SCORE, i.e., no information
beyond the robot's measurements. In comparing SCORE to Odom-R we observe that
the SCORE initialization typically obtains an average APE roughly two orders of
magnitude less than that obtained by Odom-R initialization. In fact, SCORE
appears to obtain more accurate trajectory estimates than those generated by
Odom-P, which assumes knowledge of each robot's first pose. These
observations suggest that SCORE is a more effective initialization strategy than
general odometry-based techniques (the current state-of-the-art initialization
for these problems) even when each robot's first pose is
available.

We note that several instances demonstrated substantial improvement in the
estimate by refining the SCORE initialization. In particular, we observed the
translation estimates were compressed (i.e., the estimated distance traveled was
notably less than the true value of 1 meter). Despite this compression in the
initialized translations, the final MAP estimate appeared high quality. We
hypothesize that good solutions are obtained from these compressed translations
because the SCORE initialization returns good initial estimates of the robot
orientations and beacon locations. This would agree with known results in
pose-graph SLAM \cite{carlone15icra} which, show that the quality of MAP
solutions depends more on good orientation initializations than good
translation initializations.

\subsection{When Does SCORE Return Poor Initializations?}
\label{sec:limitations}

\ScoreDeterminantPlot

Though SCORE generally performed well as an initialization technique, we
observed several multi-robot RA-SLAM instances where SCORE produced poor
initializations. These poor initializations were due to the cost function
tending towards a trivial solution of all zeros (i.e., $\rot_i = 0_{d \times d},
\tran_i = \dist_i = 0_{d}$). While the constraint $\rot_0 = I_{d}$ should
prevent this, in practice certain scenarios demonstrate the
remaining variables rapidly decaying to zero. This phenomenon is unsurprising,
as it is found in similar SLAM relaxations \cite{Rosen15icra,carlone15icra} and
the zeros elements indeed minimizes the problem cost. Importantly, we
note that these failures can be readily detected by observing the determinants
of the estimated rotations, which will quickly decay to less than
$\expnumber{5}{-2}$ (see \cref{fig:rotation-determinants}).
As a result, failures can be identified and alternative initializations used in
this case.

We find that the experiments with failed (poor) initializations resemble the
conditions identified by \cite{tseng07siam} for the SNL SOCP derived.  Namely,
the SOCP relaxation returns good solutions when the robot trajectories had
well-constrained poses (via range measurements) within the convex hull of robot
poses connected via odometry to $\rot_0$ (i.e., the trajectory of the robot with
constrained first rotation). Though at present we lack formal analysis,
empirical results suggest the analysis and connections to rigidity theory in
\cite{tseng07siam} are closely linked to SCORE.

\section{Conclusions}

We presented SCORE, a novel SOCP relaxation of the RA-SLAM problem. Through
simulated and real-world experiments we demonstrated SCORE outperforms
state-of-the-art initialization techniques and can obtain high-quality
initializations for RA-SLAM problems. The initializations provided by SCORE
resulted in up to two orders of magnitude reduction in average pose error when
compared to existing odometry-based initializations. Notably, we showed that
SCORE can determine good initializations for the challenging multi-robot RA-SLAM
problem when using only odometry and inter-robot range measurements. Along the
way, we derived a QCQP formulation of the RA-SLAM problem, which is of
independent scientific interest and connects RA-SLAM to a broader body of work.
Finally, we discussed empirically-derived hypotheses as to the conditions under
which SCORE can be expected to return a good initialization and drew connections
to existing theoretical analysis in SNL.

\bibliographystyle{IEEEtran}
\bibliography{IEEEabrv,references}

\begin{thebibliography}{10}
\providecommand{\url}[1]{#1}
\csname url@rmstyle\endcsname
\providecommand{\newblock}{\relax}
\providecommand{\bibinfo}[2]{#2}
\providecommand\BIBentrySTDinterwordspacing{\spaceskip=0pt\relax}
\providecommand\BIBentryALTinterwordstretchfactor{4}
\providecommand\BIBentryALTinterwordspacing{\spaceskip=\fontdimen2\font plus
\BIBentryALTinterwordstretchfactor\fontdimen3\font minus
  \fontdimen4\font\relax}
\providecommand\BIBforeignlanguage[2]{{%
\expandafter\ifx\csname l@#1\endcsname\relax
\typeout{** WARNING: IEEEtran.bst: No hyphenation pattern has been}%
\typeout{** loaded for the language `#1'. Using the pattern for}%
\typeout{** the default language instead.}%
\else
\language=\csname l@#1\endcsname
\fi
#2}}

\bibitem{boroson20iros}
E.~R. Boroson, R.~Hewitt, N.~Ayanian, and J.-p.~D. Croix, ``{Inter-Robot Range
  Measurements in Pose Graph Optimization},'' \emph{2020 IEEE/RSJ International
  Conference on Intelligent Robots and Systems (IROS)}, pp. 4806--4813, 2020.

\bibitem{funabiki21ral}
N.~Funabiki, B.~Morrell, J.~Nash, and A.~A. Agha-Mohammadi, ``{Range-Aided
  Pose-Graph-Based SLAM: Applications of Deployable Ranging Beacons for Unknown
  Environment Exploration},'' \emph{IEEE Robotics and Automation Letters},
  vol.~6, no.~1, pp. 48--55, 2021.

\bibitem{Newman03icra}
P.~Newman and J.~Leonard, ``Pure range-aided sub-sea {SLAM},'' in \emph{IEEE
  Intl. Conf. on Robotics and Automation (ICRA)}, vol.~2, Sept. 2003, pp.
  1921--1926.

\bibitem{Bahr06iser}
A.~Bahr and J.~Leonard, ``Cooperative {L}ocalization for {A}utonomous
  {U}nderwater {V}ehicles,'' in \emph{Intl. Sym. on Experimental Robotics
  (ISER)}, Rio de Janeiro, Brasil, July 2006.

\bibitem{Bahr12iros}
A.~Bahr, J.~Leonard, and A.~Martinoli, ``Dynamic positioning of beacon vehicles
  for coop- erative underwater navigation,'' in \emph{IEEE/RSJ Intl. Conf. on
  Intelligent Robots and Systems (IROS)}, Algarve, Portugal, 2012.

\bibitem{boyd04book}
S.~Boyd and L.~Vandenberghe, \emph{Convex optimization}.\hskip 1em plus 0.5em
  minus 0.4em\relax Cambridge university press, 2004.

\bibitem{carlone15icra}
L.~Carlone, R.~Tron, K.~Daniilidis, and F.~Dellaert, ``{Initialization
  techniques for 3D SLAM: A survey on rotation estimation and its use in pose
  graph optimization},'' \emph{Proceedings - IEEE International Conference on
  Robotics and Automation}, vol. 2015-June, no. June, pp. 4597--4604, 2015.

\bibitem{carlone15iros}
L.~Carlone, D.~M. Rosen, G.~Calafiore, J.~J. Leonard, and F.~Dellaert,
  ``Lagrangian duality in {3D} {SLAM}: Verification techniques and optimal
  solutions,'' in \emph{IEEE/RSJ Intl. Conf. on Intelligent Robots and Systems
  (IROS)}, Hamburg, Germany, sep 2015.

\bibitem{giamou19ral}
M.~Giamou, Z.~Ma, V.~Peretroukhin, and J.~Kelly, ``{Certifiably Globally
  Optimal Extrinsic Calibration From Per-Sensor Egomotion},'' \emph{IEEE
  Robotics and Automation Letters}, vol.~4, no.~2, pp. 367--374, 2019.

\bibitem{so07mathematicalprogramming}
A.~M.~C. So and Y.~Ye, ``{Theory of semidefinite programming for sensor network
  localization},'' \emph{Mathematical Programming}, vol. 109, no. 2-3, pp.
  367--384, 2007.

\bibitem{rosen19ijrr}
D.~M. Rosen, L.~Carlone, A.~S. Bandeira, and J.~J. Leonard, ``{SE}-sync: A
  certifiably correct algorithm for synchronization over the special euclidean
  group,'' \emph{The International Journal of Robotics Research}, vol.~38, no.
  2-3, pp. 95--125, 2019.

\bibitem{dellaert2017factor}
F.~Dellaert, M.~Kaess, \emph{et~al.}, ``Factor graphs for robot perception,''
  \emph{Foundations and Trends{\textregistered} in Robotics}, vol.~6, no. 1-2,
  pp. 1--139, 2017.

\bibitem{menegatti09icra}
E.~Menegatti, A.~Zanella, S.~Zilli, F.~Zorzi, and E.~Pagello, ``{Range-only
  slam with a mobile robot and a wireless sensor networks},'' \emph{Proceedings
  - IEEE International Conference on Robotics and Automation}, pp. 8--14, 2009.

\bibitem{Djugash09iros}
J.~Djugash, S.~Singh, and B.~P. Grocholsky, ``Modeling mobile robot motion with
  polar representations,'' in \emph{IEEE/RSJ Intl. Conf. on Intelligent Robots
  and Systems (IROS)}, Oct. 2009.

\bibitem{djugash09springer}
J.~Djugash and S.~Singh, ``A robust method of localization and mapping using
  only range,'' \emph{Springer Tracts in Advanced Robotics}, vol.~54, pp.
  341--351, 2009.

\bibitem{gonzalez09ras}
J.~Gonz{\'{a}}lez, J.~L. Blanco, C.~Galindo, A.~Ortiz-de Galisteo, J.~A.
  Fern{\'{a}}ndez-Madrigal, F.~A. Moreno, and J.~L. Mart{\'{i}}nez, ``{Mobile
  robot localization based on Ultra-Wide-Band ranging: A particle filter
  approach},'' \emph{Robotics and Autonomous Systems}, vol.~57, no.~5, pp.
  496--507, 2009.

\bibitem{blanco08icra}
J.~L. Blanco, J.~Gonz{\'{a}}lez, and J.~A. Fern{\'{a}}ndez-Madrigal, ``{A pure
  probabilistic approach to range-only SLAM},'' \emph{Proceedings - IEEE
  International Conference on Robotics and Automation}, pp. 1436--1441, 2008.

\bibitem{blanco08iros}
J.~L. Blanco, J.~A. Fern{\'{a}}ndez-Madrigal, and J.~Gonz{\'{a}}lez,
  ``{Efficient probabilistic range-only SLAM},'' \emph{2008 IEEE/RSJ
  International Conference on Intelligent Robots and Systems, IROS}, pp.
  1017--1022, 2008.

\bibitem{herranz14icra}
F.~Herranz, A.~Llamazares, E.~Molinos, and M.~Ocana, ``A comparison of slam
  algorithms with range only sensors,'' in \emph{IEEE Intl. Conf. on Robotics
  and Automation (ICRA)}, 2014, pp. 4606--4611.

\bibitem{Olson06joe}
E.~Olson, J.~Leonard, and S.~Teller, ``Robust range-aided beacon
  localization,'' \emph{IEEE Journal of Oceanic Engineering}, vol.~31, no.~4,
  pp. 949--958, Oct. 2006.

\bibitem{boots13icml}
B.~Boots and G.~J. Gordon, ``{A Spectral learning approach to range-only
  SLAM},'' \emph{30th International Conference on Machine Learning, ICML 2013},
  vol.~28, no. PART 1, pp. 19--26, 2013.

\bibitem{guo2017ijmav}
K.~Guo, Z.~Qiu, W.~Meng, L.~Xie, and R.~Teo, ``Ultra-wideband based cooperative
  relative localization algorithm and experiments for multiple unmanned aerial
  vehicles in gps denied environments,'' \emph{International Journal of Micro
  Air Vehicles}, vol.~9, no.~3, pp. 169--186, 2017.

\bibitem{li20arxiv}
\BIBentryALTinterwordspacing
S.~Li, M.~Coppola, C.~{De Wagter}, and G.~C. H.~E. de~Croon, ``{An autonomous
  swarm of micro flying robots with range-based relative localization},'' 2020.
  [Online]. Available: \url{http://arxiv.org/abs/2003.05853}
\BIBentrySTDinterwordspacing

\bibitem{zhou08tro}
X.~S. Zhou and S.~I. Roumeliotis, ``{Robot-to-robot relative pose estimation
  from range measurements},'' \emph{IEEE Transactions on Robotics}, vol.~24,
  no.~6, pp. 1379--1393, 2008.

\bibitem{trawny07iros}
N.~Trawny, X.~S. Zhou, K.~X. Zhou, and S.~I. Roumeliotis, ``3d relative pose
  estimation from distance-only measurments,'' in \emph{IEEE/RSJ Intl. Conf. on
  Intelligent Robots and Systems (IROS)}, 2007, pp. 1071----1078.

\bibitem{trawny10rss}
N.~Trawny, X.~S. Zhou, and S.~I. Roumeliotis, ``3d relative pose estimation
  from six distances,'' in \emph{Robotics: Science and Systems (RSS)}, vol.~5,
  2010, pp. 233--240.

\bibitem{trawny10tro}
N.~Trawny, X.~S. Zhou, K.~Zhou, and S.~I. Roumeliotis, ``Interrobot
  transformations in 3-d,'' \emph{{IEEE} Trans. Robotics}, vol.~26, no.~2, pp.
  226--243, 2010.

\bibitem{jiang20itaes}
B.~Jiang, B.~D. Anderson, and H.~Hmam, ``{3-D Relative Localization of Mobile
  Systems Using Distance-Only Measurements via Semidefinite Optimization},''
  \emph{IEEE Transactions on Aerospace and Electronic Systems}, vol.~56, no.~3,
  pp. 1903--1916, 2020.

\bibitem{li20iros}
M.~Li, G.~Liang, H.~Luo, H.~Qian, and T.~L. Lam, ``{Robot-to-robot relative
  pose estimation based on semidefinite relaxation optimization},'' \emph{IEEE
  International Conference on Intelligent Robots and Systems}, pp. 4491--4498,
  2020.

\bibitem{Li2022ral}
M.~Li, T.~L. Lam, and Z.~Sun, ``{3-D Inter-Robot Relative Localization via
  Semidefinite Optimization},'' \emph{IEEE Robotics and Automation Letters},
  vol.~7, no.~4, pp. 10\,081--10\,088, 2022.

\bibitem{martinec07cvpr}
D.~Martinec and T.~Pajdla, ``{Robust rotation and translation estimation in
  multiview reconstruction},'' \emph{Proceedings of the IEEE Computer Society
  Conference on Computer Vision and Pattern Recognition}, 2007.

\bibitem{govindu01cvpr}
V.~M. Govindu, ``Combining two-view constraints for motion estimation,''
  \emph{Proceedings of the IEEE Computer Society Conference on Computer Vision
  and Pattern Recognition}, vol.~2, pp. 218--225, 2001.

\bibitem{fredriksson13lnsc}
J.~Fredriksson and C.~Olsson, ``Simultaneous multiple rotation averaging using
  lagrangian duality,'' \emph{Lecture Notes in Computer Science (including
  subseries Lecture Notes in Artificial Intelligence and Lecture Notes in
  Bioinformatics)}, vol. 7726 LNCS, no. PART 3, pp. 245--258, 2013.

\bibitem{tron14tac}
R.~Tron and R.~Vidal, ``{Distributed 3-D localization of camera sensor networks
  from 2-D image measurements},'' \emph{IEEE Transactions on Automatic
  Control}, vol.~59, no.~12, pp. 3325--3340, 2014.

\bibitem{carlone14ijrr}
L.~Carlone, R.~Aragues, J.~A. Castellanos, and B.~Bona, ``{A fast and accurate
  approximation for planar pose graph optimization},'' \emph{International
  Journal of Robotics Research}, vol.~33, no.~7, pp. 965--987, 2014.

\bibitem{briales17ral}
J.~Briales and J.~Gonzalez-Jimenez, ``Cartan-sync: Fast and global
  se(d)-synchronization,'' \emph{IEEE Robotics and Automation Letters}, vol.~2,
  no.~4, pp. 2127--2134, 2017.

\bibitem{tron15rssworkshop}
R.~Tron, D.~M. Rosen, and L.~Carlone, ``On the inclusion of determinant
  constraints in lagrangian duality for 3d slam,'' \emph{Proc. Robotics:
  Science and Systems (RSS), Workshop "The Problem of Mobile Sensors"}, vol.~2,
  no.~1, 2015.

\bibitem{fan20tro}
\BIBentryALTinterwordspacing
T.~Fan, H.~Wang, M.~Rubenstein, and T.~Murphey, ``{CPL-SLAM: Efficient and
  Certifiably Correct Planar Graph-Based SLAM Using the Complex Number
  Representation},'' \emph{IEEE Transactions on Robotics}, vol.~36, pp.
  1719----1737, 2020. [Online]. Available:
  \url{http://arxiv.org/abs/2007.06708}
\BIBentrySTDinterwordspacing

\bibitem{briales18cvpr}
J.~Briales, L.~Kneip, and J.~Gonzalez-Jimenez, ``{A Certifiably Globally
  Optimal Solution to the Non-minimal Relative Pose Problem},'' in
  \emph{Proceedings of the IEEE Computer Society Conference on Computer Vision
  and Pattern Recognition}, 2018, pp. 145--154.

\bibitem{garcia-salguero21ivc}
M.~Garcia-Salguero, J.~Briales, and J.~Gonzalez-Jimenez, ``{Certifiable
  relative pose estimation},'' \emph{Image and Vision Computing}, vol. 109,
  2021.

\bibitem{pacholska20ral}
M.~Pacholska, F.~D{\"{u}}mbgen, and A.~Scholefield, ``{Relax and Recover:
  Guaranteed Range-Only Continuous Localization},'' \emph{IEEE Robotics and
  Automation Letters}, vol.~5, no.~2, pp. 2248--2255, 2020.

\bibitem{biswas06tase}
P.~Biswas, T.~C. Liang, K.~C. Toh, Y.~Ye, and T.~C. Wang, ``{Semidefinite
  programming approaches for sensor network localization with noisy distance
  measurements},'' \emph{IEEE Transactions on Automation Science and
  Engineering}, vol.~3, no.~4, pp. 360--371, 2006.

\bibitem{shamsi13dgo}
D.~Shamsi, N.~Taheri, Z.~Zhu, and Y.~Ye, ``Conditions for correct sensor
  network localization using sdp relaxation,'' in \emph{Discrete geometry and
  optimization}.\hskip 1em plus 0.5em minus 0.4em\relax Springer, 2013, pp.
  279--301.

\bibitem{tseng07siam}
P.~Tseng, ``{Second-order cone programming relaxation of sensor network
  localization},'' \emph{SIAM Journal on Optimization}, vol.~18, no.~1, pp.
  156--185, 2007.

\bibitem{naddafzadeh-shirazi14twc}
G.~Naddafzadeh-Shirazi, M.~B. Shenouda, and L.~Lampe, ``{Second order cone
  programming for sensor network localization with anchor position
  uncertainty},'' \emph{IEEE Transactions on Wireless Communications}, vol.~13,
  no.~2, pp. 749--763, 2014.

\bibitem{doherty01infocom}
L.~Doherty, K.~S. Pister, and L.~{El Ghaoui}, ``{Convex position estimation in
  wireless sensor networks},'' \emph{Proceedings - IEEE INFOCOM}, vol.~3, pp.
  1655--1663, 2001.

\bibitem{srirangarajan08twc}
S.~Srirangarajan, A.~Tewfik, and Z.~Q. Luo, ``{Distributed sensor network
  localization using SOCP relaxation},'' \emph{IEEE Transactions on Wireless
  Communications}, vol.~7, no.~12, pp. 4886--4895, 2008.

\bibitem{chiuso08infosys}
A.~Chiuso, G.~Picci, and S.~Soatto, ``Wide-sense estimation on the special
  orthogonal group,'' \emph{Communications in Information and Systems}, vol.~8,
  no.~3, pp. 185--200, 2008.

\bibitem{alizadeh03mathematicalprogramming}
F.~Alizadeh and D.~Goldfarb, ``{Second-order cone programming},''
  \emph{Mathematical Programming, Series B}, vol.~95, no.~1, pp. 3--51, 2003.

\bibitem{gurobi}
\BIBentryALTinterwordspacing
{Gurobi Optimization, LLC}, ``{Gurobi Optimizer Reference Manual},'' 2021.
  [Online]. Available: \url{https://www.gurobi.com}
\BIBentrySTDinterwordspacing

\bibitem{dellaert2012techreport}
F.~Dellaert, ``Factor graphs and gtsam: A hands-on introduction,'' Georgia
  Institute of Technology, Tech. Rep., 2012.

\bibitem{drake}
\BIBentryALTinterwordspacing
R.~Tedrake and the Drake Development~Team, ``Drake: Model-based design and
  verification for robotics,'' 2019. [Online]. Available:
  \url{https://drake.mit.edu}
\BIBentrySTDinterwordspacing

\bibitem{grupp2017evo}
M.~Grupp, ``evo: Python package for the evaluation of odometry and slam.''
  \url{https://github.com/MichaelGrupp/evo}, 2017.

\bibitem{Rosen15icra}
D.~Rosen, C.~DuHadway, and J.~J. Leonard, ``A convex relaxation for approximate
  global optimization in simultaneous localization and mapping,'' in \emph{IEEE
  Intl. Conf. on Robotics and Automation (ICRA)}, May 2015.

\end{thebibliography}

\end{document}